\documentclass{article}

\usepackage{arxiv}

\usepackage[utf8]{inputenc} 
\usepackage[T1]{fontenc}    
\usepackage{hyperref}       
\usepackage{url}            
\usepackage{booktabs}       
\usepackage{amsfonts}       
\usepackage{nicefrac}       
\usepackage{microtype}      
\usepackage{lipsum}		
\usepackage{graphicx}
\usepackage[numbers]{natbib}
\usepackage{doi}

\usepackage{xcolor}

\usepackage{multirow}
\usepackage{longtable}
\usepackage{tabularx}
\usepackage{tabu}
\usepackage{amsmath}
\usepackage{lscape}

\title{A Brief Review of Hypernetworks in Deep Learning}

\author{{\hspace{1mm}Vinod Kumar Chauhan$^1$\thanks{Accepted to \textbf{Artificial Intelligence Review} (Springer Nature). Corresponding author: Vinod Kumar Chauhan (vinod.kumar@eng.ox.ac.uk)},\; Jiandong Zhou$^1$,\; Ping Lu$^1$,\; Soheila Molaei$^1$\; and David A. Clifton$^{1,2}$} \\
	$^1$Institute of Biomedical Engineering,
	University of Oxford, OX3 7DQ, UK\\
        $^2$Oxford-Suzhou Institute of Advanced Research (OSCAR), Suzhou, China
}

\renewcommand{\headeright}{}
\renewcommand{\undertitle}{}

\hypersetup{
    pdftitle={A Brief Review of Hypernetworks in Deep Learning},
    pdfkeywords={Deep learning, Hypernetworks},
}

\begin{document}
\maketitle

\begin{abstract}
Hypernetworks, or hypernets for short, are neural networks that generate weights for another neural network, known as the target network. They have emerged as a powerful deep learning technique that allows for greater flexibility, adaptability, dynamism, faster training, information sharing, and model compression. Hypernets have shown promising results in a variety of deep learning problems, including continual learning, causal inference, transfer learning, weight pruning, uncertainty quantification, zero-shot learning, natural language processing, and reinforcement learning.
Despite their success across different problem settings, there is currently no comprehensive review available to inform researchers about the latest developments and to assist in utilizing hypernets. To fill this gap, we review the progress in hypernets. We present an illustrative example of training deep neural networks using hypernets and propose categorizing hypernets based on five design criteria: inputs, outputs, variability of inputs and outputs, and the architecture of hypernets. We also review applications of hypernets across different deep learning problem settings, followed by a discussion of general scenarios where hypernets can be effectively employed. Finally, we discuss the challenges and future directions that remain underexplored in the field of hypernets.
We believe that hypernetworks have the potential to revolutionize the field of deep learning. They offer a new way to design and train neural networks, and they have the potential to improve the performance of deep learning models on a variety of tasks. Through this review, we aim to inspire further advancements in deep learning through hypernetworks.
\end{abstract}


\section{Introduction}
\label{sec_intro}
Deep learning has revolutionized the field of artificial intelligence by enabling remarkable advancements in various domains, including computer vision \cite{chauhan2024hcr}, natural language processing \cite{devlin-etal-2019-bert}, causal inference \cite{chauhan2023adversarial}, and reinforcement learning \cite{li2017deep}. Standard deep neural networks (DNNs) have proven to be powerful tools for learning complex representations from data. However, despite their success, standard DNNs remain restrictive in certain conditions. For example, once a DNN is trained, its weights as well as its architecture are fixed \cite{rohanian-etal-2023-using,vaswani2017attention}, and any changes to weights or architecture require re-training the DNN. This lack of adaptability and dynamism restricts the flexibility of DNNs, making them less suitable for scenarios where dynamic adjustments or data adaptivity are required \cite{ha2017hypernetworks,brock2018smash}. DNNs generally have a large number of weights and need substantial amounts of data to optimize those weights \cite{alzubaidi2021review}. This can be challenging in situations where large amounts of data are not available. For example, in healthcare, collecting sufficient data for rare diseases can be particularly difficult due to the limited number of patients available per year \cite{wiens2014study}. Finally, uncertainty quantification in DNNs' predictions is essential as it provides a measure of confidence, enabling better decision-making in high-stakes applications \cite{chauhan2024continuous}. Existing uncertainty quantification techniques have limitations, such as the need to train multiple models \cite{abdar2021review}, and uncertainty quantification is still considered an open problem \cite{kristiadi2019predictive}. Similarly, domain adaptation, domain generalization, adversarial defence, neural style transfer, and neural architecture search are important problems that remain unsolved, where hypernets can provide effective solutions as discussed in Section~\ref{sec_applications}.

Hypernetworks (or hypernets in short) have emerged as a promising architectural paradigm to enhance the flexibility (through data adaptivity and dynamic architectures) and performance of DNNs. Hypernets are a class of neural networks that generate the weights/parameters of another neural network called the target/main/primary network, where both networks are trained in an end-to-end differentiable manner \cite{ha2017hypernetworks}. Hypernets complement existing DNNs and provide a new framework to train DNNs, resulting in a new class of DNNs called HyperDNNs (please refer to Section~\ref{sec_background} for details). The key characteristics and advantages of hypernets that offer applications across different problem settings are discussed below.

\begin{enumerate}
    \item[(a)] Soft weight sharing: Hypernetworks can be trained to generate the weights of multiple DNNs for solving related tasks \cite{chauhan2024dynamic,Oswald2020Continual}. This is called soft weight sharing because, unlike hard weight sharing which involves shared layers among tasks (e.g., in multitasking), different DNNs are generated by a common hypernet through task conditioning. This helps share information among tasks and can be used for transfer learning or dynamic information sharing \cite{chauhan2024dynamic}.
        
    \item[(b)] Dynamic architectures: Hypernetworks can be used to generate the weights of a network with a dynamic architecture, where the number of layers or the structure of the network changes during training or inference. This can be particularly useful for tasks where the target network structure is not known at training time \cite{ha2017hypernetworks}.
        
    \item[(c)] Data-adaptive DNNs: Unlike standard DNNs whose weights are fixed at inference time, HyperDNNs can generate a target network customized to the needs of the data. In such cases, hypernets are conditioned on the input data to adapt to the data \cite{sun2017hypernetworks}.
    
    \item[(d)] Uncertainty quantification: Hypernets can effectively train uncertainty-aware DNNs by leveraging techniques like sampling multiple inputs from the noise distribution \cite{krueger2018bayesian} or incorporating dropout within the hypernets themselves \cite{chauhan2023dynamic}. By generating multiple sets of weights for the main network, hypernets create an ensemble of models, each with different parameter configurations. This ensemble-based approach aids in estimating uncertainty in the model predictions, a crucial aspect for safety-critical applications like healthcare, where having a measure of confidence in predictions is essential.
    
    \item[(e)] Parameter efficiency: HyperDNNs, i.e., DNNs trained with hypernets, can have fewer weights than the corresponding standard DNNs, resulting in weight compression \cite{zhao2020meta}. This can be particularly useful when working with limited resources, limited data, or high-dimensional data and can result in faster training than the corresponding DNN \cite{navon2021learning}.
\end{enumerate}

Ha~et.~al~\cite{ha2017hypernetworks} coined the term hypernets (also referred to as meta-networks or meta-models) and trained the target network and hypernet in an end-to-end differentiable way. However, the concept of learnable context-dependent weights was discussed even earlier, such as \textit{fast weights} in \cite{schmidhuber1992learning,schmidhuber1993self} and HyperNEAT \cite{stanley2009hypercube}. Our discussion on hypernets focuses on neural networks generating weights for the target neural network due to their popularity, expressiveness, and flexibility \cite{vaswani2017attention,chauhan2024hcr}.
Recently, hypernets have gained significant attention and have produced state-of-the-art (SOTA) results across several deep learning problems, including ensemble learning \cite{kristiadi2019predictive}, multitasking \cite{tay2021hypergrid}, neural architecture search \cite{zhang2018graph}, continual learning \cite{Oswald2020Continual}, weight pruning \cite{liu2019metapruning}, Bayesian neural networks \cite{deutsch2019generative}, generative models \cite{deutsch2019generative}, hyperparameter optimization \cite{lorraine2018stochastic}, information sharing \cite{chauhan2024dynamic}, adversarial defence \cite{sun2017hypernetworks}, and reinforcement learning (RL) \cite{rezaei2023hypernetworks} (please refer to Section~\ref{sec_applications} for more details).

Despite the success of hypernets across different problem settings, to the best of our knowledge, there is no review of hypernets to guide researchers about the developments and to help in utilizing hypernets. To fill this gap, we provide a brief review of hypernets in deep learning. We illustrate hypernets using an example and differentiate HyperDNNs from DNNs (Section~\ref{sec_background}). To facilitate better understanding and organization, we propose a systematic categorization of hypernets based on five distinct design criteria, resulting in different classifications that consider factors such as (i) input characteristics, (ii) output characteristics, (iii) variability of inputs, (iv) variability of outputs, and (v) the architecture of hypernets (Section~\ref{sec_categorization}). Furthermore, we offer a comprehensive overview of the diverse applications of hypernets in deep learning, spanning various problem settings (Section~\ref{sec_applications}). By examining real-world applications, we aim to demonstrate the practical advantages and potential impact of hypernetworks. Additionally, we discuss some scenarios and pose direct questions to understand if we can apply hypernets to a given problem (Section~\ref{sec_when}). Finally, we discuss the challenges and future directions of hypernet research (Section~\ref{sec_challenges}). This includes addressing initialization, stability, and complexity concerns, as well as exploring avenues for enhancing the theoretical understanding and uncertainty quantification of DNNs. By providing a comprehensive review of hypernetworks, this paper aims to serve as a valuable resource for researchers and practitioners in the field. Through this review, we hope to inspire further advancements in deep learning by leveraging the potential of hypernets to develop more flexible, high-performing models.

\textbf{Contributions}: This review paper makes the following key contributions:
\begin{itemize}
    \item To the best of our knowledge, we present the first review on hypernetworks in deep learning, which have shown impressive results across several deep learning problems.

    \item We propose categorizing hypernets based on five design criteria, leading to different classifications of hypernets, such as based on inputs, outputs, variability of inputs and outputs, and architecture of hypernets.
    
    \item We present a comprehensive overview of applications of hypernetworks across different problem settings, such as uncertainty quantification, continual learning, causal inference, transfer learning, and federated learning, and summarize our review, as per our categorization, in a table (Table~\ref{tab_hypernets}).

    \item We explore broad scenarios for hypernet applications, drawing from existing use cases and hypernet characteristics. This exploration aims to equip researchers with actionable insights into when to leverage hypernets in their problem setting.

    \item Finally, we identify the challenges and future directions of hypernetwork research, including initialization, stability, scalability, and efficiency concerns, and the need for theoretical understanding and interpretability of hypernetworks. By highlighting these areas, we aim to inspire further advancements in hypernetworks and provide guidance for researchers interested in addressing these challenges.
\end{itemize}

The rest of the paper is organized as follows: Section~\ref{sec_background} provides a comprehensive background on hypernets, while Section~\ref{sec_categorization} introduces a novel categorization scheme for hypernets. The diverse applications of hypernets across various problems are discussed in Section~\ref{sec_applications}, followed by an exploration of specific scenarios where hypernets can be effectively employed in Section~\ref{sec_when}. Addressing challenges and delineating future research directions is the focus of Section~\ref{sec_challenges}, and finally, the concluding remarks are discussed in Section~\ref{sec_conclusion}.

\section{Background}
\label{sec_background}
\begin{figure}[htb!]
    \centering
    \includegraphics[width=0.95\textwidth]{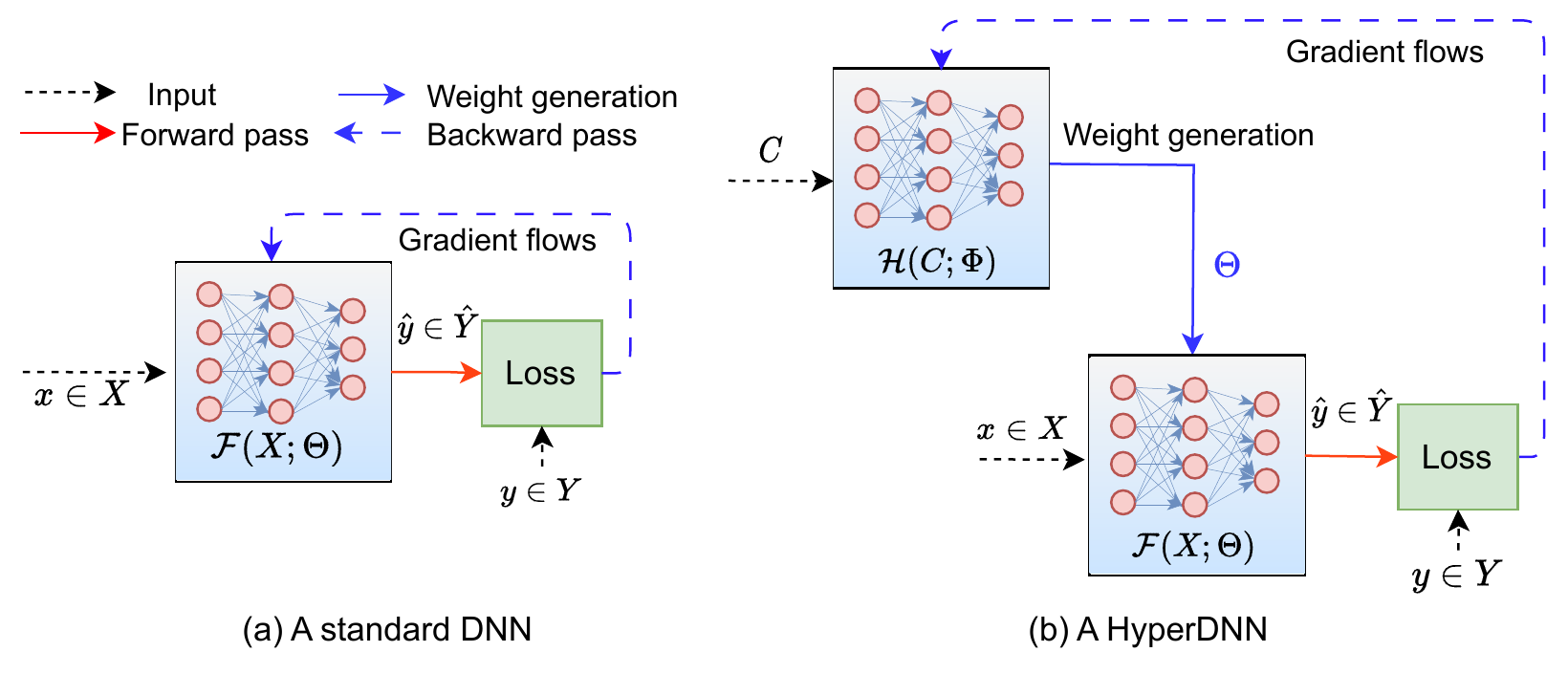}
    \caption{An overview of the architectures and gradient flows for a standard DNN $\mathcal{F}(X; \Theta)$ and the same DNN implemented with hypernets, referred to as HyperDNN $\mathcal{F}(X; \Theta)=\mathcal{F}(X; \mathcal{H}(C; \Phi))$. For the DNN, gradients flow through the DNN, and DNN weights $\Theta$ are learned during training. For the HyperDNN, gradients flow through the hypernet, and hypernet weights $\Phi$ are learned during training to produce DNN weights $\Theta$ as outputs.}
    \label{fig_hypernet_example}
\end{figure}

In this section, we discuss and differentiate the workings of standard deep neural networks (DNNs) and DNNs trained with hypernetworks, referred to as HyperDNNs, using a generic example. Fig.~\ref{fig_hypernet_example} illustrates the structural differences and gradient flows in DNNs and HyperDNNs. Both solve the same problem using the same DNN architecture at inference time. However, differences exist in their training processes, specifically in gradient flow and weight optimization, making hypernets an alternative way of training DNNs.

Let us denote a dataset using ${X,Y}$ to solve a general task $\mathcal{T}$, where $X$ is a matrix of features and $Y$ is a vector of labels, and $x\in X$ denotes one data point and $y\in Y$ is the corresponding label. Let a DNN be denoted as a function $\mathcal{F}(X; \Theta)$, where $X$ denotes the inputs and $\Theta$ represents the weights of the DNN. During the forward pass, inputs $x\in X$ pass through the layers of $\mathcal{F}$ to produce predictions $\hat{y} \in \hat{Y}$, which are then used along with true labels $ y \in Y$ to calculate an objective function that measures the discrepancy between actual values and the values predicted by the model using a loss function $\mathcal{L}(Y, \hat{Y})$. During the backward pass, DNNs typically use backpropagation to propagate the error backwards through the layers and calculate gradients of $\mathcal{L}$ with respect to $\Theta$. Optimization algorithms, such as Adam \cite{kingma2014adam}, use these gradients to update the weights. At the end of the training, we receive optimized weights $\Theta$ that are used at inference time in the DNN $\mathcal{F}(X; \Theta)$ to make predictions with the test data for solving task $\mathcal{T}$. Thus, in standard DNNs, $\Theta$ are the learnable weights.

Hypernets provide an alternative way of learning weights $\Theta$ of the DNN $\mathcal{F}(X; \Theta)$ to solve task $\mathcal{T}$, where $\Theta$ are not directly learned but are generated by another neural network. In this framework, we solve the same task using the same DNN architecture but with a different training approach. Let a hypernet be denoted as $\mathcal{H}(C; \Phi)$ which generates the task-specific weights of the DNN $\mathcal{F}(X; \Theta)$, where $C$ is a task-specific context vector that acts as input to $\mathcal{H}$ and $\Phi$ are weights of the hypernet $\mathcal{H}$. That is, $\Theta = \mathcal{H}(C; \Phi)$ where $\Phi$ are the only learnable weights in the overall architecture. The context vector $C$ can be generated from the data \cite{Alaluf2022HyperStyle}, sampled from a noise distribution \cite{krueger2018bayesian}, or correspond to task identity/embedding \cite{armstrong2021continual}. During the forward pass, a task-specific context vector $C$ is passed to the hypernet $\mathcal{H}$ which generates weights $\Theta$ for the DNN $\mathcal{F}$. Then, like a standard DNN, an input $x\in X$ is passed through the DNN $\mathcal{F}$ to predict the output $Y$, and the loss is calculated as $\mathcal{L}(Y, \hat{Y})$. However, during the backward pass, the error is backpropagated through the hypernet $\mathcal{H}$ and gradients of $\mathcal{L}$ are calculated with respect to the weights of the hypernet $\Phi$. The learning algorithm optimizes $\Phi$ to generate $\Theta$ so that performance on the target task $\mathcal{T}$ is optimized. At test time, $\Theta$ generated from the optimized hypernet $\mathcal{H}$ are used in the DNN $\mathcal{F}(X; \Theta)$ to make predictions with the test data for solving task $\mathcal{T}$. The optimization problems for the standard DNN and the HyperDNN can be written as follows (ignoring regularization terms for simplicity):

\begin{equation}
    \text{DNN:} \quad \min_{\Theta}\;\; \mathcal{F}(X; \Theta), \qquad \text{HyperDNN:} \quad \min_{\Phi}\;\; \mathcal{F}(X; \Theta)=\mathcal{F}(X; \mathcal{H}(C; \Phi)).
\end{equation}
Thus, DNNs learn their weights\footnote{we have used weights and parameters interchangeably} directly from the data, while in HyperDNNs the weights of the hypernet are learned, and the weights of the DNN are generated by the hypernet. For a specific example of a comparison of DNN and HyperDNN architectures and their workings, please refer to our work in causal inference \cite{chauhan2024dynamic}.

As discussed in Section~\ref{sec_intro}, training a DNN with a hypernet, i.e., HyperDNN presents several advantages over directly training a DNN. However, these advantages are application-specific and cannot be generalized across all tasks or applications. For instance, a key feature of hypernets is soft-weight sharing, which enables information sharing among related components. This information sharing is particularly valuable in settings with limited data, leading to performance improvements for HyperDNNs in such scenarios. In general, HyperDNNs are beneficial for applications with limited data, problems requiring data-adaptive networks, dynamic network architectures, parameter efficiency, and uncertainty quantification. A detailed discussion of scenarios where HyperDNNs can be useful is provided in Section~\ref{sec_when}.

In general, if a task can be solved using standard DNNs, it is advisable to use them instead of hypernets. As depicted in Figure~\ref{fig_hypernet_example}, HyperDNNs require an additional DNN to solve the same task. Despite the advantages offered by hypernets, this additional DNN introduces complexities in training and implementing HyperDNNs. For example, the initialization of HyperDNNs is more challenging than DNNs because the weights of the target network are generated at the output layer of the hypernet. Classical initialization techniques do not guarantee that the weights of the target network are initialized within the same range. However, adaptive optimizers, such as Adam \cite{kingma2014adam}, can mitigate this issue to some extent.
Another significant challenge with HyperDNNs is their scalability. Since the weights of the target network are generated at the output layer of the hypernet, this approach can present difficulties when dealing with large target networks. Scalability issues can be managed using various weight generation strategies. Therefore, when using HyperDNNs, practitioners should consider employing adaptive optimizers, implementing different weight generation strategies, and using approaches to stabilize training, such as spectral norms. For a detailed discussion on the challenges associated with HyperDNNs, please refer to Section~\ref{sec_challenges}.

\section{Categorization of Hypernetworks}
\label{sec_categorization}
In this section, we propose to categorize the hypernetworks based on five design criteria, as depicted in Fig.~\ref{fig_hypernet_categorization} and as given below: 
\begin{enumerate}
    \item[(a)] Input-based, i.e., what kind of input is taken by the hypernetworks to generate the target neural network weights? 
    \item[(b)] Output-based, i.e., how are the outputs, that is, the target weights generated? 
    \item[(c)] Variability of inputs, i.e., are the inputs of hypernet fixed?
    \item[(d)] Variability of outputs, i.e., does the target network have a fixed number of weights? and 
    \item[(e)] Architecture-based, i.e., what kind of architecture does hypernet use to generate the target weights?
\end{enumerate}
We discuss these in the following subsections. One can categorize hypernets based on the architecture of the target network but that is not considered because hypernets mostly generate target weights independent of their architecture.

\begin{figure}[htb!]
    \centering
    \includegraphics[width=0.99\textwidth]{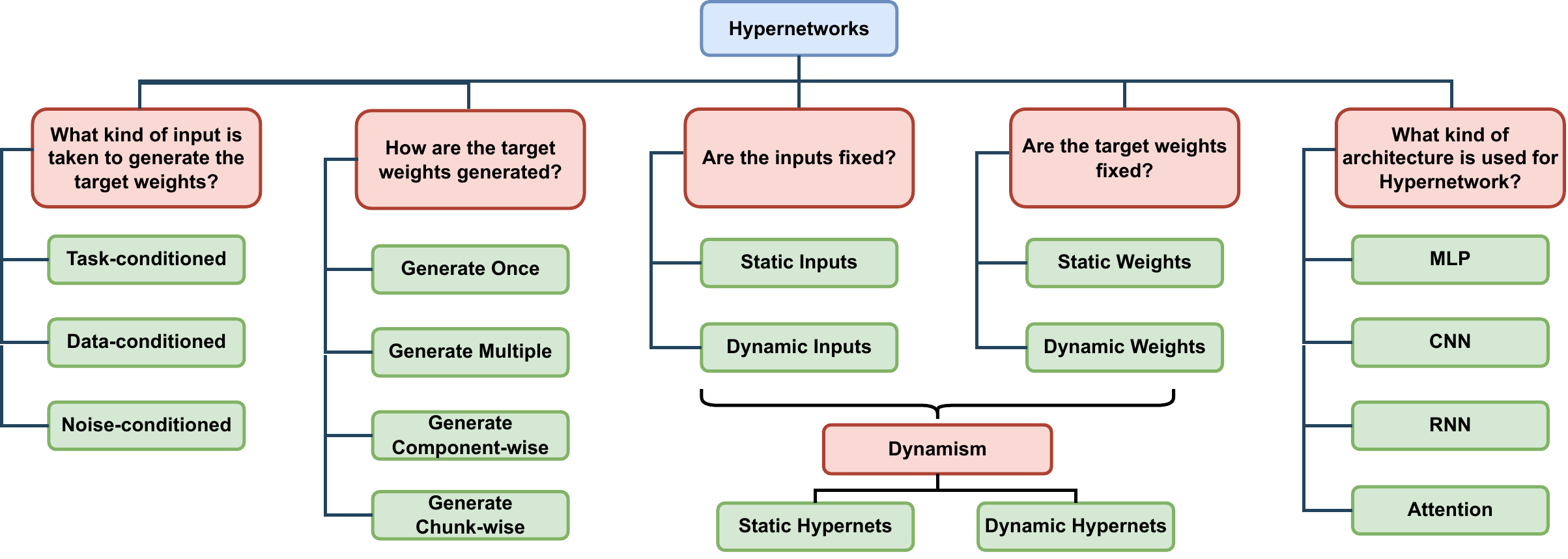}
    \caption{Proposed categorization of hypernets based on five design criteria.}
    \label{fig_hypernet_categorization}
\end{figure}

\subsection{Input-based Hypernetworks}
\label{subsec_input_based}
Hypernetworks take a context vector as an input and generate weights of the target DNN as output. Depending on what context vector is used, we can have the following types of hypernetworks.

\textbf{Task-conditioned hypernetworks}: These hypernetworks take task-specific information as input. The task information can be in the form of task identity/embedding, hyperparameters, architectures, or any other task-specific cues. The hypernetwork generates weights that are tailored to the specific task. This allows the hypernet to adapt its behavior accordingly and allows information sharing, through soft weight sharing of hypernets, among the tasks, resulting in better performance on the tasks. For example, \citet{chauhan2024dynamic} applied hypernets to solve treatment effects estimation problem in causal inference that uses an identity or embedding of potential outcome (PO) functions to generate weights corresponding to the PO function. The hypernetworks enabled dynamic end-to-end inter-treatment information sharing among treatment groups and helped to calculate reliable treatment estimates in observational studies with limited-size datasets. Similarly, task-conditioned hypernets have been used to solve other problems, including multitasking \cite{navon2021learning}, natural language processing (NLP) \cite{ha2017hypernetworks}, and continual learning \cite{Oswald2020Continual}.

\textbf{Data-conditioned hypernetworks}: These hypernetworks are conditioned on the data that the target network is being trained on. The hypernetwork generates weights based on the characteristics of the input data. This enables the neural network to dynamically adjust its behavior based on the specific input pattern or features, leading to more flexible and adaptive models, and resulting in better generalization to unseen data. For example, \citet{Alaluf2022HyperStyle} applied hypernets for image editing where the input of hypernet is based on the input images and initial approximation of reconstruction to generate modulations to the weights of the pre-trained generator. Similarly, data-conditioned hypernets have been used to solve other problems, such as adversarial defence \cite{sun2017hypernetworks}, knowledge graphs learning \cite{balavzevic2019hypernetwork} and shape learning \cite{littwin2019deep}.

\textbf{Noise-conditioned hypernetworks}: These hypernetworks are not conditioned on any input data or task cues, but rather on randomly sampled noise. This makes them more general-purpose and helps in predictive uncertainty quantification for DNNs, but it also means that they may not perform as well as task-conditioned or data-conditioned hypernetworks on multiple tasks or datasets. For example, \citet{krueger2018bayesian} applied hypernetworks to approximate Bayesian inference in the DNNs and evaluated the approach for active learning, model uncertainty, regularization, and anomaly detection. Similarly, noise-conditioned hypernets have been used to solve other problems, such as manifold learning \cite{deutsch2019generative} and uncertainty quantification \cite{ratzlaff2019hypergan}.

These different types of conditioning enable hypernetworks to enhance the flexibility (through adaptability and dynamic architectures), and performance of deep learning models in various contexts. {\color{black}The specific type of hypernetwork that is used will depend on the specific task or application. For example, task-conditioned hypernets are suitable for information sharing among multiple tasks, data-conditioned hypernets are suitable to deal with conditions where DNN need to adapt to input data, and noise-conditioned hypernets are suitable for uncertainty quantification in the predictions.}

\subsection{Output-based Hypernetworks}
\label{subsec_output_based}
Based on the outputs of hypernets, i.e., weight generation strategy, we classify hypernetworks according to whether all weights are generated together or not. This classification of hypernetworks is important because it controls the scalability and complexity of the hypernetworks, as typically DNNs have a large number of weights, and producing all of them together can make the size of the last layer of hypernets large. So, there are ways to manage the complexity of the hypernets that lead to different strategies of weight generation, as discussed below. It is possible to train HyperDNN with fewer weights than the target DNN -- this is called weight compression \cite{zhao2020meta}. {\color{black}We compared and summarized the characteristics of various weight generation strategies in Table~\ref{tab_output_hypernets}. The first column represents the considered characteristic for comparison, while the following three columns correspond to three different weight generation strategies. The values in each row indicate whether a particular weight generation strategy provides the specified feature or not.}

\textbf{Generate Once}:
These hypernetworks generate weights of the entire target DNN altogether. This approach uses all the generated weights, and weights of each layer are generated together, unlike the other weight generation strategies. However, this weight generation approach is not suitable for large target networks because that can lead to complex hypernets. For example, \citet{shamsian2021personalized,Galanti2020,zhang2018graph} used generate once weight generation.

\textbf{Generate Multiple}: These hypernetworks have multiple heads for producing weights (sometimes referred to as split/multi-head hypernets) and this weight generation approach can complement the other approaches. This simplifies the complexity and reduces the number of weights required in the last layer of the hypernets by the number of head times. This approach does not need additional embeddings, and in general, uses all the generated weights, unlike component-wise and chunk-wise weight generation approaches where some weights remain unused. For example, \citet{beck2023hypernetworks,rezaei2023hypernetworks,chauhan2024dynamic} used generate multiple strategy to produce target weights.

\textbf{Generate Chunk-wise}: Chunk-wise hypernetworks generate weights of the target network in chunks. This can lead to not using some of the generated weights because the weights are generated as per the chunk size, which may not match the layer sizes. If the chunk size is smaller than the layer size, then all the weights of a layer may not be generated together. Moreover, these hypernets need additional embeddings to distinguish different chunks and to produce specific weights for the chunks. However, overall chunk-wise weight generation leads to reducing complexity and improving the scalability of hypernets. For example, \citet{chauhan2024dynamic,Oswald2020Continual} used chunk-wise weight generation.

\textbf{Generate Component-wise}: Component-wise weights generation strategy generates weights for each individual component (such as layer or channel) of the target model separately. This is helpful in generating specific weights because different layers or channels represent different features or patterns in the network. However, similar to the chunk-wise approach, component-wise hypernets need an embedding for each component to distinguish among different components and produce weights specific to that component. They also help to reduce the complexity and improve the scalability of hypernets. Since the weights are generated as per the size of the largest layer so this weight generation approach can lead to not using some of weights in smaller layers. This strategy can be seen as a special case of a chunk-wise weight generation approach, where one chunk is equal to the size of one component. For example, \citet{zhao2020meta,Alaluf2022HyperStyle,mahabadi2021parameter} used component-wise weight generation.

By classifying hypernetworks based on their weight generation strategy, we can make informed choices that may help control the scalability and complexity of the hypernetworks effectively. Each type of weight generation strategy offers unique benefits and considerations based on the specific characteristics and requirements of the task at hand. The comparative study of characteristics of different weight generation approaches is summarized in Table~\ref{tab_output_hypernets}.

{\color{black}
\begin{table}[htb!]
\caption{Comparison of different weight generation strategies, i.e., output-based hypernetworks.}
\label{tab_output_hypernets}
    \begin{tabular}{p{0.20\textwidth}p{0.15\textwidth}p{0.18\textwidth}p{0.18\textwidth}p{0.18\textwidth}} \\ \hline
    \textbf{Weight-generation$\rightarrow$ /Characteristics$\downarrow$} &
      \textbf{Generate Once} &
      \textbf{Generate Component-wise} &
      \textbf{Generate Chunk-wise} &
      \textbf{Generate Multiple} \\ \hline
    Weight generation &
      Generates all target weights together &
      Generates target weights for one component at a time &
      Generates target weights in chunks &
      Complements all other weight generation strategies so can generate weights like any of the other \\ \hline
    Efficient use of generated weights &
      Yes &
      No as some weights can stay unused &
      Yes &
      Depends on the base strategy \\ \hline
    Are all weights of a layer generated together &
      Yes &
      Yes &
      No &
      Depends on the base strategy \\  \hline
    The complexity of output space &
      Highest &
      Lower than generate once &
      Lowest &
      Can further improve Chunk-wise generation \\  \hline
    The complexity of input space &
      Lowest &
      More complex than `generate once' but lower than chunk-wise generation if the number of target layers is fewer than the number of chunks &
      Highest (assuming the number of chunks is more than the number of target layers) &
      Does not have any effect on input space complexity\\ 
      \hline
    \end{tabular}
\end{table}
}

\subsection{Variability of Inputs}
\label{subsec_input_variability}
We can categorize hypernets based on the variability of the inputs. We have two classes, static inputs and dynamic inputs, as discussed below.

{\color{black}\textbf{Static Inputs:}} If the inputs are predefined and are fixed then the hypernet is called static with respect to the inputs. For example, multitasking \cite{mahabadi2021parameter} has fixed number of tasks leading to fixed number of inputs. It is to be noted that here fixed input only means fixed tasks identities, however hypernets can learn embeddings for different tasks.

{\color{black}\textbf{Dynamic Inputs:} If the inputs change and generally are dependent on data on which the target network is trained, then the hypernet is called dynamic with respect to the inputs.} Dynamic inputs help hypernetworks to introduce a new level of adaptability by dynamically generating the weights of the target network. This dynamic weight generation enables hypernetworks to respond to input-dependent context and adjust their behavior accordingly. By generating network weights based on specific inputs, hypernetworks can capture intricate patterns and dependencies that may vary across different instances of data. This adaptability leads to enhanced model performance, especially in scenarios with complex and evolving data distributions \cite{volk2022example}. Thus, dynamic input-based hypernets help in domain adaptation \cite{volk2022example}, density estimation \cite{Hofer2023} and knowledge graph learning \cite{balavzevic2019hypernetwork} etc.

This can be seen as a super categorization over input-based hypernets where task-conditioned hypernets fall in the static inputs category while random-noise and data-conditioned hypernets fall in the dynamic category. Both the categories have their own advantages as static inputs help in information sharing \cite{chauhan2024dynamic}, transfer learning \cite{Oswald2020Continual}, and are suitable where we have multiple tasks to solve \cite{shamsian2021personalized}. On the other hand, dynamic inputs give hypernets adaptability to new conditions unknown during training \cite{balavzevic2019hypernetwork}.

\subsection{Variability of Outputs}
\label{subsec_output_variability}
When classifying hypernetworks based on the nature of the target network's weights, we can categorize them into two types, static outputs or dynamic outputs, as discussed below.

{\color{black}\textbf{Static Outputs:} If weights of the target network are fixed in size, then the hypernet is called static with respect to the outputs.} In this case, the target network is also static. For example, \citet{pan2018hyperst,szatkowski2022hypersound} produce static weights.

{\color{black}\textbf{Dynamic Outputs:} If weights of the target network are not fixed, i.e., the architecture varies in size, then the hypernet is called dynamic with respect to the outputs, and the target network is also a dynamic network as it can have different architecture depending on the input of the hypernet.} The dynamic weights can be generated, mainly, in two situations, first when the hypernet architecture is dynamic, e.g., \citet{ha2017hypernetworks} used recurrent neural network (RNN) to propose HyperRNN based on non-shared weights. Second, the dynamic weights can be generated when the inputs are dynamic, i.e., hypernet adapts as per the input data, e.g., \citet{littwin2019deep} applied convolutional neural network (CNN) based hypernet to generate dynamic weights for shape learning from an image of a shape. Similarly, \citet{peng2020cream,li2020dhp} also produce dynamic weights.

{\color{black}
\subsection{Dynamism in Hypernetworks}
\label{subsec_dynamism}
This is a super categorization of Subsection~\ref{subsec_input_variability} and \ref{subsec_output_variability} into broader category based on the dynamism in inputs or outputs of the hypernets, as discussed below.

\textbf{Static Hypernets:} If input of a hypernet is fixed, i.e., predefined and number of weights produced by hypernet for the target network are fixed, i.e., the architecture is fixed, then the hypernet is called as a static hypernet. This kind of hypernets work with predefined inputs, e.g., task identities, which can be learned as embeddings, but the tasks being solved remain same. For example, heterogeneous treatment effect estimation \cite{chauhan2024dynamic} where number of treatment groups or potential outcome functions are fixed, and architecture of the target network (in this case potential outcome functions) is also fixed.

\textbf{Dynamic Hypernets:} If input of a hypernet is based on input of target network, i.e., input data, or number of weights produced by hypernet for the target network are variable, i.e., the architecture is dynamic, then the hypernet is called as a dynamic hypernet. For example, \citet{sendera2023hypershot} applied data-conditioned hypernet to few-shot learning by combining kernels and hypernets. The kernels were used to extract support information from data of different tasks that act as input to the hypernet which generates weights for the target task. \citet{zhang2018graph} applied hypernetworks for neural architecture search where they modeled neural architectures of a DNN as graph and used them as input to hypernet to generate the target network weights. So, the target network has variable architecture, and is a dynamic hypernet based on the dynamic outputs.
}

\subsection{Architecture of Hypernetworks}
\label{subsec_architecture_based}
In the categorization of hypernetworks based on their architectures, we can classify them into four major types: multi-layer perceptrons (MLPs), convolutional neural networks (CNNs), recurrent neural networks (RNNs), and attention-based networks, as given below.

{\color{black}\textbf{MLPs:}} MLP based hypernetworks employ a dense and fully connected architecture, allowing every input neuron to connect with every output neuron. This architecture enables a comprehensive weight generation process by considering the entire input information, e.g., \cite{chauhan2024dynamic}.

{\color{black}\textbf{CNNs:}} CNN hypernetworks, on the other hand, leverage convolutional layers to capture local patterns and spatial information. These hypernetworks excel in tasks involving spatial data, such as an image or video analysis, by extracting features from the input and generating weights or parameters accordingly, e.g., \citet{nirkin2021hyperseg} employed MLP to implement hypernets.

{\color{black}\textbf{RNNs:}} RNN hypernetworks incorporate recurrent connections in their architecture, facilitating feedback loops and sequential information processing. They dynamically generate weights or parameters based on previous states or inputs, making them well-suited for tasks involving sequential data, such as natural language processing or time series analysis, e.g., \citet{ha2017hypernetworks} employed RNN to implement hypernets.

{\color{black}\textbf{Attention}} Attention-based hypernetworks incorporate attention mechanisms \cite{vaswani2017attention} into their architecture. By selectively focusing on relevant input features, these hypernetworks generate weights for the target network, allowing them to capture long-range dependencies and improve the quality of generated outputs, e.g., \citet{volk2022example} employed attention to implement hypernets.

Each type of architecture has its own strengths and applicability, enabling hypernetworks to adapt and generate weights in a manner that aligns with the specific characteristics and demands of the target network and the data being processed.

\section{Applications of Hypernetworks}
\label{sec_applications}
Hypernetworks have demonstrated their effectiveness and versatility across a wide range of domains and tasks in deep learning. In this section, we discuss some of the important applications\footnote{{\color{black}We have explored 50 important papers (arranged by publication year) while considering at least one application in each distinct problem setting. This is not an exhaustive list and it is possible that we may have missed important references.}} of hypernetworks and highlight their contributions to advancing the SOTA in these areas. We summarize the applications of hypernets as per our proposed categorization and also provide links to code repositories for the benefit of the researchers, wherever available, in Table~\ref{tab_hypernets}.

\textbf{Continual Learning}:
Continual learning, also known as lifelong learning or incremental learning, is a machine learning paradigm that focuses on the ability of a model to learn and adapt continuously over time, in a sequential manner, without forgetting previously learned knowledge. Unlike traditional batch learning, which assumes static and independent training and testing sets, continual learning deals with dynamic and non-stationary data distributions, where new data arrives incrementally, and the model needs to adapt to these changes while retaining previously acquired knowledge. The challenge in continual learning lies in mitigating \textit{catastrophic forgetting}, which refers to the tendency of a model to forget previously learned information when it is trained on new data. To address this, various strategies have been proposed, including regularization techniques, rehearsal methods, dynamic architectures, and parameter isolation. {\color{black}\citet{Oswald2020Continual} modeled each incrementally obtained dataset as a task and applied task-conditioned hypernets for continual learning -- this helped to share information among tasks. To address the catastrophic forgetting issue, they proposed a regularizer for rehearsing task-specific weight realizations rather than the data from previous tasks.} They achieved SOTA results on benchmarks and empirically showed that the task-conditioned hypernets have a long capacity to retain memories of previous tasks. {\color{black}Similarly, \citet{huang2021continual,ehret2021continual} applied task-conditioned hypernets to continual learning in reinforcement learning (RL).}

\textbf{Federated Learning}:
Federated Learning is a decentralized approach to machine learning where the training process is distributed across multiple devices or edge devices, without the need to centralize data in a single location. In this paradigm, each device or edge node locally trains a model using its own data, and only the model updates, rather than the raw data, are shared and aggregated on a central server. This enables collaborative learning while preserving data privacy and security. It also reduces communication costs and latency, making it suitable for scenarios with limited bandwidth or intermittent connectivity. {\color{black}\citet{shamsian2021personalized} modeled each client machine as a task and applied task-conditioned hypernets to federated learning problem. They trained a central hypernet to generate the weights for the client models. This allowed information sharing across different clients while making the hypernet size independent of communication cost, as hypernet weights are never transmitted.} The hypernet-based federated learning achieved the SOTA results and also showed better generalization to new clients whose distributions were different than the existing clients. {\color{black}\citet{litany2022federated} extended this work to heterogeneous clients, i.e., clients with different neural architectures, using graph hypernetworks \cite{zhang2018graph}.}

{\scriptsize
\LTcapwidth=\linewidth
\begin{longtable}[htb!]{rrp{3cm}lllllllllllp{1.5cm}l}
\caption{{\color{black}Important applications of hypernetworks, arranged by ascending publication year,} and their categorization based on Input: (i) task-conditioned, (ii) noise-conditioned, and (iii) data-conditioned; output, i.e., weight generation: (i) generate once, (ii) generate component-wise, (iii) generate chunk-wise, and (iv) generate multiple; Input variability: (i) static inputs, and (ii) dynamic inputs; Output variability: (i) static weights, and (ii) dynamic weights; and architecture of hypernets (SN: Serial Number, Ref.: Reference, DL: Deep Learning, RL: Reinforcement learning).}
\label{tab_hypernets}\\
\hline
\multicolumn{1}{c}{\multirow{2}{*}{\textbf{SN}}} &
  \multicolumn{1}{c}{\multirow{2}{*}{\textbf{Ref.}}} &
  \multicolumn{1}{c}{\multirow{2}{*}{\textbf{DL Problem}}} &
  \multicolumn{3}{c}{\textbf{Input}} &
  \multicolumn{4}{c}{\textbf{Output}} &
  \multicolumn{2}{c}{\textbf{Input Var.}} &
  \multicolumn{2}{c}{\textbf{Out Var.}} &
  \multicolumn{1}{c}{\multirow{2}{*}{\textbf{Architecture}}} &
  \multicolumn{1}{c}{\multirow{2}{*}{\textbf{Code}}} \\
  \multicolumn{1}{c}{} &
  \multicolumn{1}{c}{} &
  \multicolumn{1}{c}{} &
  \multicolumn{1}{c}{(i)} &
  \multicolumn{1}{c}{(ii)} &
  \multicolumn{1}{c}{(iii)} &
  \multicolumn{1}{c}{(i)} &
  \multicolumn{1}{c}{(ii)} &
  \multicolumn{1}{c}{(iii)} &
  \multicolumn{1}{c}{(iv)} &
  \multicolumn{1}{c}{(i)} &
  \multicolumn{1}{c}{(ii)} &
  \multicolumn{1}{c}{(i)} &
  \multicolumn{1}{c}{(ii)} &
  \multicolumn{1}{c}{} &
  \multicolumn{1}{c}{}                                     \\
  \hline
\endhead
\endfoot
\endlastfoot
1  & \cite{ha2017hypernetworks}        & Image classification, NLP                                           & \checkmark &            &            &            & \checkmark &            &            &                   & \checkmark        & \checkmark        & \checkmark        & RNN, MLP                 &                                                                           \\
2  & \cite{krueger2018bayesian}        & Uncertainty quantification                                                                  &            & \checkmark &            & \checkmark &            &            &            &                   & \checkmark        & \checkmark        &                   & MLP                      &                                                                           \\
3  & \cite{sun2017hypernetworks}       & Adversarial defence                                                 &            &            & \checkmark &            & \checkmark &            &            &                   & \checkmark        &                   & \checkmark        & MLP                      &                                                                           \\
4  & \cite{lorraine2018stochastic}     & Hyperparameter optimization                                         & \checkmark &            &            & \checkmark &            &            &            &                   & \checkmark        & \checkmark        &                   & MLP                      &                                                                           \\
5  & \cite{brock2018smash}             & Neural architecture search                                          & \checkmark &            &            &            &            & \checkmark &            &                   & \checkmark        &                   & \checkmark        & CNN                      & \href{https://github.com/ajbrock/SMASH}{Link}                             \\
6  & \cite{pan2018hyperst}             & Spatio-temporal learning                                           &            &            & \checkmark & \checkmark &            &            &            &                   & \checkmark        & \checkmark        &                   & MLP, RNN, CNN            &                                                                           \\
7  & \cite{zhang2018graph}             & Neural architecture search                                          & \checkmark &            &            & \checkmark &            &            &            &                   & \checkmark        & \checkmark        &                   & MLP                      &                                                                           \\
8  & \cite{deutsch2019generative}      & Manifold learning                                                   &            & \checkmark &            &            & \checkmark &            &            &                   & \checkmark        &                   & \checkmark        & CNN                      &                                                                           \\
9  & \cite{ratzlaff2019hypergan}       & Uncertainty quantification                                                                  &            & \checkmark &            &            & \checkmark &            & \checkmark &                   & \checkmark        & \checkmark        &                   & GAN                      &                                                                           \\
10 & \cite{liu2019metapruning}         & Weight pruning                                                      & \checkmark &            &            & \checkmark &            &            &            &                   & \checkmark        & \checkmark        &                   & MLP                      & \href{https://github.com/liuzechun/MetaPruning}{Link}                     \\
11 & \cite{balavzevic2019hypernetwork} & Knowledge graphs learning                                           &            &            & \checkmark &            & \checkmark &            &            &                   & \checkmark        & \checkmark        &                   & MLP                      & \href{https://github.com/ibalazevic/HypER}{Link}                          \\
12 & \cite{littwin2019deep}            & Shape learning                                                      &            &            & \checkmark &            &            &            & \checkmark &                   & \checkmark        &                   & \checkmark        & CNN                      & \href{https: //github.com/gidilittwin/Deep-Meta}{Link}                    \\
13 & \cite{kristiadi2019predictive}    & Uncertainty quantification                                                                  &            &            & \checkmark & \checkmark &            &            &            &                   & \checkmark        & \checkmark        &                   & MLP                      &                                                                           \\
14 & \cite{klocek2019hypernetwork}     & Image processing                                                    &            &            & \checkmark &            &            &            & \checkmark &                   & \checkmark        & \checkmark        &                   & CNN                      &                                                                           \\
15 & \cite{Oswald2020Continual}        & Continual learning, transfer learning                               & \checkmark &            &            & \checkmark & \checkmark & \checkmark &            & \checkmark        &                   & \checkmark        &                   & MLP                      & \href{https://github.com/chrhenning/hypercl}{Link}                        \\
16 & \cite{zhao2020meta}               & Few-shot learning                                                   & \checkmark &            &            &            &            & \checkmark &            & \checkmark        &                   & \checkmark        &                   & MLP                      &                                                                           \\
17 & \cite{Galanti2020}                & Complexity of NN                                                    &            &            & \checkmark & \checkmark &            &            &            &                   & \checkmark        & \checkmark        &                   & MLP                      &                                                                           \\
18 & \cite{li2020dhp}                  & Weight pruning                                                      & \checkmark &            &            &            & \checkmark &            &            &                   & \checkmark        &                   & \checkmark        & MLP                      & \href{https://github.com/ofsoundof/dhp}{Link}                             \\
19 & \cite{peng2020cream}              & Neural architecture search                                          & \checkmark &            &            &            & \checkmark &            &            &                   & \checkmark        &                   & \checkmark        & CNN                      & \href{https://github.com/microsoft/Cream}{Link}                           \\
20 & \cite{navon2021learning}          & Pareto-Front Learning (multi-tasking, fairness, image segmentation) & \checkmark &            &            &            &            &            & \checkmark & \checkmark        &                   & \checkmark        &                   & MLP                      & \href{https://github.com/AvivNavon/pareto-hypernetworks}{Link}            \\
21 & \cite{shamsian2021personalized}   & Federated Learning                                                  & \checkmark &            &            & \checkmark &            &            &            & \checkmark        &                   & \checkmark        &                   & MLP                      & \href{https://github.com/AvivSham/pFedHN}{Link}                           \\
22 & \cite{nirkin2021hyperseg}         & Semantic segmentation                                               &            &            & \checkmark &            & \checkmark &            & \checkmark &                   & \checkmark        &                   & \checkmark        & CNN                      & \href{https://nirkin.com/hyperseg}{Link}                                  \\
23 & \cite{mahabadi2021parameter}      & Multitasking, NLP, language model                                   & \checkmark &            &            &            & \checkmark &            &            & \checkmark        &                   & \checkmark        &                   & MLP                      & \href{https://github.com/rabeehk/hyperformer}{Link}                       \\
24 & \cite{sarafian2021recomposing}    & RL                                              & \checkmark &            &            & \checkmark &            &            &            & \checkmark        &                   & \checkmark        &                   & MLP                      & \href{https://github.com/keynans/HypeRL}{Link}                            \\
25 & \cite{huang2021continual}         & Continual RL                                                        & \checkmark &            &            & \checkmark &            &            &            & \checkmark        &                   & \checkmark        &                   & MLP                      & \href{https://github.com/rvl-lab-utoronto/HyperCRL}{Link}                 \\
26 & \cite{Shih2021}                   & Density estimation                                                  & \checkmark &            &            &            & \checkmark &            &            & \checkmark        &                   & \checkmark        &                   & MLP                      &                                                                           \\
27 & \cite{Muller2021}                 & Neural image enhancement                                            &            &            & \checkmark & \checkmark &            &            &            &                   & \checkmark        &                   & \checkmark        & MLP                      &                                                                           \\
28 & \cite{henning2021posterior}       & Continual learning                                                  & \checkmark &            &            & \checkmark &            &            &            & \checkmark        &                   & \checkmark        &                   & MLP                      &                                                                           \\
29 & \cite{ehret2021continual}         & Continual learning                                                  & \checkmark &            &            &            &            & \checkmark &            & \checkmark        &                   & \checkmark        &                   & MLP                      & \href{https://github.com/mariacer/cl_in_rnns}{Link}                       \\
30 & \cite{lamb2021contextual}         & Adaptation of neural network architectures                          &            &            & \checkmark & \checkmark &            &            &            &                   & \checkmark        &                   & \checkmark        & MLP                      &                                                                           \\
31 & \cite{Nguyen2021}                 & Network compression                                                 & \checkmark &            &            & \checkmark &            &            &            &                   & \checkmark        & \checkmark        &                   & MLP                      &                                                                           \\
32 & \cite{Bensadoun2021}              & Internal learning (computer vision)                                 &            &            & \checkmark &            &            &            & \checkmark &                   & \checkmark        & \checkmark        &                   & CNN                      & \href{https://github.com/RaphaelBensTAU/MetaInternalLearning}{Link}       \\
33 & \cite{belbute-peres2021hyperpinn} & Learning differential equations                                     & \checkmark &            &            & \checkmark &            &            &            &                   & \checkmark        & \checkmark        &                   & MLP                      &                                                                           \\
34 & \cite{Alaluf2022HyperStyle}       & Image editing                                                       &            &            & \checkmark &            & \checkmark &            &            &                   & \checkmark        & \checkmark        &                   & CNN                      & \href{https://yuval-alaluf.github.io/hyperstyle/}{Link}                   \\
35 & \cite{volk2022example}            & Domain adaptation, NLP                                              & \checkmark &            & \checkmark & \checkmark &            &            &            &                   & \checkmark        & \checkmark        &                   & Attention              & \href{https://github.com/TomerVolk/Hyper-PADA}{Link}                      \\
36 & \cite{oh2022cvae}                 & Autonomous driving                                                  &            &            & \checkmark &            &            &            & \checkmark &                   & \checkmark        &                   & \checkmark        & Attention, RNN, CNN, MLP &                                                                           \\
37 & \cite{wullach2022character}       & NLP                                                                 & \checkmark &            & \checkmark &            & \checkmark &            &            & \checkmark        &                   & \checkmark        & \checkmark        & MLP                      & \href{https://github.com/tomerwul/CharLevelHyperNetworks}{Link}           \\
38 & \cite{qu2022hmoe}                 & Domain generalization                                               & \checkmark &            &            & \checkmark &            &            &            & \checkmark        &                   & \checkmark        &                   & MLP                      &                                                                           \\
39 & \cite{Spurek2022}                 & 3D point cloud processing                                           &            &            & \checkmark & \checkmark &            &            &            &                   & \checkmark        & \checkmark        &                   & MLP                      &                                                                           \\
40 & \cite{dinh2022hyperinverter}      & Image processing                                                    &            &            & \checkmark & \checkmark &            &            &            &                   & \checkmark        & \checkmark        &                   & CNN                      & \href{https://github.com/VinAIResearch/HyperInverter}{Link}               \\
41 & \cite{yin2022sylph}               & Few-shot learning                                                   &            &            & \checkmark & \checkmark &            &            &            &                   & \checkmark        & \checkmark        &                   & CNN                      & \href{https://github.com/facebookresearch/sylph-few-shot-detection}{Link} \\
42 & \cite{chauhan2024dynamic}         & Treatment effects estimation                                    & \checkmark &            &            & \checkmark & \checkmark & \checkmark & \checkmark & \checkmark        &                   & \checkmark        &                   & MLP                      &                                                                           \\
43 & \cite{beck2023hypernetworks}      & Meta-RL                                         & \checkmark &            &            &            &            &            & \checkmark & \checkmark        &                   & \checkmark        &                   & MLP                      &                                                                           \\
44 & \cite{rezaei2023hypernetworks}    & Zero-shot RL                                                        & \checkmark &            &            &            &            &            & \checkmark &                   & \checkmark        & \checkmark        &                   & MLP                      & \href{https://sites.google.com/view/hyperzero-rl}{Link}                   \\
45 & \cite{szatkowski2022hypersound}   & Sound representation                                                &            &            & \checkmark & \checkmark &            &            &            &                   & \checkmark        & \checkmark        &                   & CNN                      &                                                                           \\
46 & \cite{Hofer2023}                  & Density estimation                                                  &            &            & \checkmark & \checkmark &            &            &            &                   & \checkmark        & \checkmark        &                   & CNN                      &                                                                           \\
47 & \cite{carrasquilla2023quantum}    & Quantum computing                                                   & \checkmark &            &            & \checkmark &            &            &            &                   & \checkmark        & \checkmark        &                   & MLP                      & \href{https://github.com/carrasqu/binncode}{Link}                         \\
48 & \cite{Ruta2023}                   & Neural style transfer                                               &            &            & \checkmark & \checkmark &            &            &            &                   & \checkmark        &                   & \checkmark        & MLP                      &                                                                           \\
49 & \cite{ferens2023hyperpose}        & Camera pose localization                                            &            &            & \checkmark &            & \checkmark &            &            &                   & \checkmark        & \checkmark        &                   & Attention, MLP           & \href{https://anonymous.4open.science/r/hyperpose-2023/README.md}{Link}   \\
50 & \cite{wu2023hyperinr}             & Knowledge distillation, visualization                               & \checkmark &            &            & \checkmark &            &            &            &                   & \checkmark        & \checkmark        &                   & MLP                      &                                                                          
\\  \hline
\end{longtable}
}

\textbf{Few-shot Learning}:
Few-shot learning is a sub-field of machine learning that focuses on training models to learn new concepts or tasks with only a limited number of training examples. Unlike traditional machine learning approaches that typically require large amounts of labeled data for each task, few-shot learning aims to generalize knowledge from a small support set of labeled examples to classify or recognize new instances. To address the practical difficulties of existing techniques to operate in high-dimensional parameter spaces with extremely limited-data settings, {\color{black}\citet{rusu2018metalearning} applied data-conditioned hypernets. They employed encoder-decoder based hypernet which learns a data-dependent latent generative representation of model parameters that shares information between different tasks through soft weight sharing of hypernets. They also achieved SOTA results and showed that the proposed technique can capture uncertainty in the data. \citet{sendera2023hypershot} also applied data-conditioned hypernet to few-shot learning by combining kernels and hypernets. The kernels were used to extract support information from data of different tasks that act as input to the hypernet which generates weights for the target task. Similarly, \citet{zhao2020meta,przewikezlikowski2022hypermaml,sendera2023general} also applied hypernets, and utilized soft weight sharing, for few-shot learning.}

\textbf{Manifold Learning}:
Manifold learning is a sub-field of machine learning that focuses on capturing the underlying structure or geometry of high-dimensional data in lower-dimensional representations or manifolds. It aims to uncover the intrinsic relationships and patterns within the data by mapping it to a lower-dimensional space, enabling better visualization, clustering, or classification. Hypernetworks can be utilized in the context of manifold learning to enhance the representation learning process. By generating weights or parameters for the target network based on the input, hypernetworks can adaptively learn a manifold that captures the intricate data structure \cite{shamsian2021personalized}. {\color{black}\citet{deutsch2019generative} applied noise-conditioned hypernetworks to map latent vectors for generating target network weights that generalize mode connectivity in loss landscape to higher dimensional manifolds.}

\textbf{AutoML}:
AutoML, short for Automated Machine Learning, refers to the development of algorithms, systems, and tools that automate various aspects of the machine learning pipeline, e.g., neural architecture search (NAS) and automated hyperparameter optimization. {\color{black}\citet{zhang2018graph} applied hypernetworks for NAS where they modeled neural architectures of a DNN as graph and used them as input to hypernet to generate the target network weights. They achieved about 10 times faster results than the SOTA. Similarly, \citet{brock2018smash,peng2020cream} present another example of application of hypernets to NAS, where they exploit soft weight sharing property of hypernets for information sharing among different architectures.} For hyperparameter optimization, {\color{black}\citet{lorraine2018stochastic} applied hypernets that take hyperparameters of the target network as input and generate optimal weights for the target network}, and hence perform joint training for target network parameters and hyperparameters which are otherwise trained in nested optimization loops. The authors proved the efficacy of the proposed technique against the SOTA to train thousands of hyperparameters.

\textbf{Pareto-front Learning}:
Pareto-front learning, also known as multi-objective optimization, is a technique that addresses problems with multiple conflicting objectives, e.g., multitasking has multiple tasks that may have conflicting gradients. It aims to find a set of solutions that represent the trade-off among different objectives, rather than a single optimal solution. In Pareto-front learning, the goal is to identify a set of solutions that cannot be improved in one objective without sacrificing performance in another objective. These solutions are referred to as Pareto-optimal or non-dominated solutions and lie on the Pareto-front, which represents the best possible trade-off between objectives. \citet{navon2021learning} applied hypernets to learn the entire Pareto-front, which at inference time takes a preferential point on the Pareto-front and generates Pareto-front weights for the target network whose loss vector is in the direction of the ray. They showed that the proposed hypernets are computationally very efficient as compared with the SOTA and can scale to large models, such as ResNet18. {\color{black}This work is further extended in \citet{hoang2023improving}, where hypernet generates multiple solutions, and \citet{tran2023framework}, which consider completed scalarization functions in the Pareto-front learning.}

\textbf{Domain Adaptation}:
Domain adaptation refers to the process of adapting a machine learning model trained on a source domain to perform well in a different target domain. It is a crucial challenge in machine learning when there is a shift or discrepancy between the distribution of the source and the target data. {\color{black}Hypernets can play a valuable role in domain adaptation by dynamically generating or adapting model parameters, architectures, or other components to effectively handle domain shifts.} For example, \citet{volk2022example} were the first to propose hypernets for domain adaptation. They used data-conditioned hypernets where examples from the target domains are used as input to hypernet that generates weights for the target network. {\color{black}This gives hypernets ability to learn and share information from existing domains with target domain through shared training}.

\textbf{Causal Inference}:
Causal inference is a field of study that focuses on understanding and estimating causal relationships between variables. It aims to uncover the cause-and-effect relationships within a system by leveraging observational or experimental data. Causal inference is particularly important when inferring the impact of treatments/ interventions/ policies on outcomes of interest. Recently, we were the first to apply hypernets to heterogeneous treatment effects (HTE) estimation problem \cite{chauhan2024dynamic}. {\color{black}We applied task-conditioned hypernets where each potential outcome (PO) function is considered as a task. Embeddings of PO functions are used as input to hypernet that generates parameters for the corresponding PO function, i.e., factual and counterfactual models. Based on soft weight sharing of hypernets, this work presents the first general mechanism to train HTE learners that enables end-to-end inter-treatment information sharing among the PO functions and helps to get reliable estimates, especially with limited-size observational data. The proposed framework also incorporates dropout in the hypernet that allows to generate multiple sets of parameters for the PO functions and helps in uncertainty quantification.}

\textbf{Uncertainty Quantification}:
Uncertainty quantification is a critical aspect of deep learning and decision-making that involves estimating and understanding the uncertainty associated with model predictions or outcomes. It provides a measure of confidence or reliability in the predictions made by a model, particularly in situations where the model encounters unseen or uncertain data. {\color{black}Hypernets can effectively train uncertainty aware DNNs by leveraging techniques like sampling multiple inputs from the noise distribution \cite{krueger2018bayesian} or incorporating dropout within the hypernets themselves \cite{chauhan2023dynamic}. By generating multiple sets of weights for the main network, hypernets create an ensemble of models, each with different parameter configurations. This ensemble-based approach aids in estimating uncertainty in the model predictions.} \citet{krueger2018bayesian} proposed Bayesian hypernets that take random noise as input to produce distributions over the weights of the target network and showed competitive performance for uncertainty. \citet{ratzlaff2019hypergan} also applied noise-conditioned hypernets for uncertainty quantification and showed that the proposed technique provides a better estimate of uncertainty as compared to the ensemble learning technique. In addition, \citet{chauhan2023dynamic} used dropout in the task-conditioned hypernets to generate multiple sets of weights for the target network and thus helping to estimate uncertainty.

\textbf{Adversarial Defence}:
Adversarial defence in deep learning refers to the techniques used to enhance the robustness and resilience of models against adversarial attacks. Adversarial attacks involve making carefully crafted perturbations to input data in order to deceive or mislead deep learning models \cite{madry2017towards}. {\color{black}By incorporating hypernetworks, models can enhance their ability to detect and defend against adversarial attacks by dynamically generating or adapting their weights or architectures.} For example, \citet{sun2017hypernetworks} generated data-dependent adaptive convolution kernels to improve the robustness of CNNs against adversarial attacks and were successful in spontaneously detecting attacks generated by Gaussian noise, fast gradient sign methods, and black-box attack methods. The models developed with hypernets are highly adaptive and customized to the data. Similarly, \citet{kristiadi2019predictive,ratzlaff2019hypergan,krueger2018bayesian} also found noise-conditioned hypernets robust to adversarial examples as compared with the SOTA.

\textbf{Multitasking}:
Multitasking refers to the capability of a model to perform multiple tasks or learn multiple objectives simultaneously. It involves leveraging shared representations and parameters across different tasks to enhance learning efficiency and overall performance. {\color{black}Hypernets can be applied in the context of multitasking to facilitate the joint learning of multiple tasks by dynamically generating or adapting the model's parameters or architectures. Specifically, we can train task-conditioned hypernets for multitasking where embedding of a task act as input to the hypernet that generates weights for the corresponding task. We can either generate entire model for each of the tasks or can only generate non-shared parts of a multitasking network. The hypernets facilitate such models to share information across different tasks as well as have specific personalized model for each task}. For example, \citet{mahabadi2021parameter} applied task-conditioned hypernets that share knowledge across the tasks as well as generate task-specific models and achieved benchmark results. \citet{navon2021learning} also studied task-conditioned hypernets for Pareto-front learning to address the conflicting gradients among different objectives and obtained impressive results on multitasking, including fairness and image segmentation.

\textbf{Reinforcement Learning}:
Reinforcement Learning (RL) focuses on training agents to make sequential decisions in an environment to maximize a cumulative reward. RL operates through an interaction loop where the agent takes actions, receives feedback in the form of rewards, and learns optimal policies through trial and error. {\color{black}Hypernets can be used to dynamically generate or adapt network architectures, model parameters, or exploration strategies in RL agents. By using a hypernetwork, the RL agent can effectively learn to customize its internal representations or policies based on the specific characteristics of the environment or task.} For example, \citet{sarafian2021recomposing} applied hypernets to generate the building blocks of RL, i.e., policy networks and Q-functions, rather than using MLPs. They showed faster training and improved performance on different algorithms for RL and in meta-RL. Similarly, noise-conditioned hypernets are used in \cite{vincent2023parameterized} to generate weights of each Bellman iteration with HyperRNN, and task-conditioned hypernets were used in RL for generalization across tasks \cite{beck2023hypernetworks}, continual RL \cite{huang2021continual}, and zero-shot learning \cite{rezaei2023hypernetworks}.

\textbf{Natural Language Processing}:
Natural language processing (NLP) is a sub-field of artificial intelligence that focuses on the interaction between computers and human language. It involves various tasks, such as language generation, sentiment analysis, machine translation, and question answering, among others. In the context of NLP, hypernets can be used to generate or adapt neural network architectures, tuning hyperparameters, for neural architecture search, and for transfer learning and domain adaptation etc. For example, {\color{black}\citet{volk2022example} applied data-conditioned hypernet for out-of-distribution (OOD) generalization. They used T5 encoder-decoder framework to generate a unique signature for each example from different source domains. This signature acts as input to the hypernet and generates parameters for the target network -- a dynamic and adaptive network. As discussed above, \citet{mahabadi2021parameter} applied task-conditioned hypernets to fine-tune the pre-trained language models by generating weights for the bottleneck adapters. In the multitasking setting, they modeled task, adapter location and layer id as different tasks and used embedding of these tasks as input to the hypernet that helps in shared learning and achieving parameter efficiency.}

\textbf{Computer Vision}:
Computer vision focuses on enabling computers to understand and interpret visual information from images or videos. Computer vision algorithms aim to replicate human visual perception by detecting and recognizing objects, understanding their spatial relationships, extracting features, and making sense of the visual scene. Some applications of hypernets in computer vision are: \citet{ha2017hypernetworks}, in their pioneering work, first applied task-conditioned hypernets for image classification, \citet{Alaluf2022HyperStyle,Muller2021} applied data-conditioned hypernets, where image acts as input to hypernet, for image enhancement, and \citet{ratzlaff2019hypergan} applied noise-conditioned hypernets for image classification. Data-conditioned hypernets are also applied to semantic segmentation in \cite{nirkin2021hyperseg}. {\color{black}Some other applications of hypernets in computer vision are camera pose estimation \cite{ferens2023hyperpose}, neural style transfer \cite{Ruta2023}, image processing/editing \cite{Alaluf2022HyperStyle}, and neural image enhancement \cite{Muller2021}. It is to be noted that computer vision is a vast subject and encompasses many problem settings discussed earlier so they can be used as such with change of domain related data or models. For example, hypernets developed for AutoML, domain adaption, continual learning, and federated learning etc. can be applied to computer vision problems as well.}

{\color{black}The above applications of hypernets are not exhaustive and some other interesting areas where hypernets have produced the SOTA results are knowledge graph learning \cite{balavzevic2019hypernetwork}, shape learning \cite{littwin2019deep}, network compression \cite{Nguyen2021}, learning differential equations \cite{belbute-peres2021hyperpinn}, 3D point cloud processing \cite{spurek2020hypernetwork}, speech processing \cite{szatkowski2022hypersound}, quantum computing \cite{carrasquilla2023quantum}, and knowledge distillation \cite{wu2023hyperinr} etc.} These applications demonstrate the wide-ranging potential of hypernetworks in deep learning, enabling adaptive and task-specific parameter generation for improved model performance and generalization.

{\color{black}
\section{When can we use Hypernets?}
\label{sec_when}
After discussing what a hypernet is, how it works, its different types, and its current applications, the most important question is when and where to utilize hypernets. This will help researchers and practitioners fully harness the benefits of this versatile technique in deep learning. One straightforward answer to the question, `When can we use Hypernets?' is `in all those application areas where it is already applied'. There is a long list of application areas where hypernets are already in use, and the reader's area of interest is likely covered. Based on the characteristics and applications of hypernets discussed above, we have generalized and formulated some questions/scenarios for readers to check if hypernets can be applied to a specific area/problem setting. If our answer is yes to any of the scenarios, then we can apply hypernets to the problem setting under consideration.

\textbf{Are there any related components in the problem setting under consideration?}\\
Here, a component can refer to a task, dataset, or neural network. This is one of the most important scenarios/questions, and several applications, as discussed above, fall under this scenario. If the answer to this question is yes, then we can employ task-conditioned hypernets to solve the problem under consideration, where task identity is used to generate the target network for the component. By conditioning on the component (task, dataset, or network), we can perform joint training of different components by exploiting the soft weight sharing of hypernets. This enables the hypernets to share information among components, leading to improved performance \cite{chauhan2024dynamic}. Thus, sharing information is the key to achieving better results for related components. The question can be reformulated as, `Do we need information sharing in our problem setting?'. All the task-conditioned applications of hypernets discussed in Table~\ref{tab_hypernets} fall under this scenario. For example, multitasking \cite{mahabadi2021parameter} has related tasks (as components), and hypernets help in shared learning while having personalized networks for each task. Similarly, continual learning \cite{Oswald2020Continual}, federated learning \cite{shamsian2021personalized}, heterogeneous treatment effects estimation \cite{chauhan2024dynamic}, transfer learning \cite{Oswald2020Continual}, and domain adaptation \cite{volk2022example} fall under this scenario.

\textbf{Do we need a data-adaptive neural network?}\\
This is another important scenario with several applications across different problem settings. In other words, we can ask, `Are we working in a setting where the target network has to be customized to the input data?' or `Are the data changing regularly?'. In this scenario, we can employ data-conditioned hypernets that take data as input and adaptively generate the parameters of the target network. During training, the hypernet takes the available data and learns the intrinsic characteristics of the data to generate the target network. Then, at inference time, it can take new data with slightly different characteristics and generate the target network based on the learned characteristics of the existing data. It is noted that there is some similarity between task-conditioned and data-conditioned settings, so some problems may be modelled using either technique. From existing research, it is unclear when to model a problem as data-conditioned or task-conditioned, and it needs to be explored. However, it will depend on the problem under consideration, the availability of data, and the number of tasks. All the data-conditioned applications of hypernets discussed in Table~\ref{tab_hypernets} fall under this scenario. For example, in neural image enhancement \cite{Muller2021}, we are interested in improving the quality of an image, so we need a target network specific to the image for a good quality output. Thus, data-conditioned hypernets are suitable for this application. Similarly, adversarial defence \cite{sun2017hypernetworks}, shape learning \cite{littwin2019deep}, camera pose estimation \cite{ferens2023hyperpose}, neural style transfer \cite{Ruta2023}, few-shot learning \cite{yin2022sylph}, and 3D point cloud processing \cite{Spurek2022} fall under this scenario.

\textbf{Do we need a dynamic neural network architecture?}\\
Here, dynamic neural network architecture means the architecture of the target network is not known or fixed at training time. This scenario has limited but important applications. In this case, a hypernet takes some information about the architecture of the target network and generates the parameters accordingly. For example, neural architecture search \cite{zhang2018graph} is such an application, which uses graph hypernetworks that take the computation graph of the target network as input to generate the network parameters. Similarly, another example of this scenario is when recurrent neural networks are implemented with hypernets \cite{ha2017hypernetworks}, which need a dynamic network architecture to account for a variable number of time-steps.

\textbf{Do we need faster training/parameter efficiency?}\
As discussed earlier, hypernets can achieve parameter efficiency or weight compression, which means that the `learnable' weights of HyperDNN are fewer than the corresponding DNN. This is expected to achieve faster training as well. This could be useful for limited resource settings and would depend on the problem setting as well as the architecture of the hypernets. For example, as discussed earlier, \citet{mahabadi2021parameter} applied task-conditioned hypernets to fine-tune pre-trained language models by generating weights for the bottleneck adapters. In the multitasking setting, they modelled task, adapter location, and layer identity as different tasks and used embeddings of these tasks as input to the hypernet that helps in shared learning and achieved parameter efficiency. Similarly, \citet{zhao2020meta} also demonstrated parameter efficiency in a few-shot learning setting.

\textbf{Do we need uncertainty quantification?}\
This is a specific application scenario for hypernets. Hypernets can be used for uncertainty quantification either using noise-conditioned hypernets \cite{krueger2018bayesian} or by using dropout in the hypernets \cite{chauhan2023dynamic}. As discussed earlier, in some settings, hypernets can produce better uncertainty estimates, e.g., \cite{krueger2018bayesian,ratzlaff2019hypergan}. However, if uncertainty estimation is the sole purpose of the study, then existing uncertainty estimation techniques must be explored first. However, using dropout \cite{srivastava2014dropout} in the hypernet architecture, similar to using dropout in standard DNNs, can complement the existing hypernets and help in uncertainty quantification.

The scenarios discussed have overlaps, so multiple scenarios can fit a problem under consideration. For example, \citet{mahabadi2021parameter} considered fine-tuning language models using hypernets, which achieved parameter efficiency and used task-conditioning (related component setting) to solve multiple tasks. Thus, by thinking about these broad scenarios, one can determine if hypernets apply to a problem setting under consideration.
}

\section{Challenges and Future Directions}
\label{sec_challenges}
Hypernetworks have shown enormous potential in enhancing deep learning models with increased flexibility, efficiency, and generalization. However, several challenges and opportunities for future research and development remain under-explored. In this section, we discuss some of the key challenges and propose potential directions for future exploration.

\textbf{Initialization Challenge}:
The initialization challenge in hypernetworks refers to the difficulty of initializing the hypernetwork parameters effectively, as finding suitable initial values for the hypernetwork parameters is far from being resolved. {\color{black}One reason for the initialization challenge is that the weights of the target network are generated at the output layer of hypernet, and weights generation does not consider layer-wise architecture of the target network. So, initialization of hypernet weights using classical initialization techniques, such as Xavier \cite{glorot2010understanding} and Kaiming initialization \cite{he2015delving}, does not guarantee that weights of target network are initialized in the same range.} The performance of the hypernetwork is highly influenced by the initial state of the target network and its parameters that are generated at the output layer of the hypernet. If the target network is poorly initialized, it can propagate errors or uncertainties to the hypernetwork, affecting its ability to generate or adapt parameters effectively. \citet{chang2020Principled} were the first to discuss the challenge of initializing hypernets. They showed that classical techniques of initializing DNNs do not work well with hypernets, however, adaptive optimizers, such as Adam \cite{kingma2014adam}, can address the issue to some extent. The authors suggested initializing the hypernet weights in a manner that allows the target network weights to approximate the conventional initialization of DNNs. However, it is difficult to adopt this because the weights of the target network are typically generated together. We may solve this challenge if weight generation process is aware of the layer-wise architecture of the target network. Moreover, recently, \citet{beck2023hypernetworks} also showed that initialization challenge of hypernets occurs even in meta-RL and classical initialization techniques fail.

\textbf{Complexity/Scalability}:
One of the primary challenges in hypernetworks is scalability and efficiency of hypernetwork-based models. As the size and complexity of target DNNs increase, hypernetworks also become very complex, e.g., the size of the output layer is typically $m\times n$ where $m$ is the number of neurons in the penultimate layer of hypernet and $n$ is the number of weights in the target network. So, hypernets may not be suitable for large models unless appropriate weight-generation strategies are developed and used. Although, there are some approaches, such as multiple weight generation \cite{chauhan2024dynamic} and chunk-wise weight generation \cite{brock2018smash} to manage the complexity of hypernets but it needs more research to address the scalability challenge and make hypernetworks more practical for real-world applications.

\textbf{Numerical Stability}:
Numerical stability in hypernetworks refers to the ability of the model to maintain accurate and reliable computations throughout the training and inference process. Hypernets, like standard neural networks, can encounter numerical stability issues \cite{sarafian2021recomposing}. One common numerical stability issue in hypernetworks is the vanishing or exploding gradients problem. During the training process, gradients can become extremely small or large, making it difficult for the model to effectively update the parameters. This can result in slow convergence or unstable training dynamics. To address numerical stability issues in hypernets, various techniques can be employed, such as careful initialization of the model's parameters, the use of gradient clipping, which bounds the gradient values to prevent them from becoming too large, and different regularization techniques such as weight decay, dropout, and spectral norm \cite{chauhan2024dynamic} that help improve numerical stability by preventing overfitting and promoting smoother optimization. Furthermore, similar to standard DNNs, using appropriate activation functions, such as ReLU or Leaky ReLU, can help alleviate the vanishing gradient problem by providing non-linearities that allow for more effective gradient propagation. It is also important to choose appropriate optimization algorithms that are known for their stability, such as Adam \cite{kingma2014adam}, which can handle the training dynamics of hypernetworks more effectively \cite{chang2020Principled}.

\textbf{Theoretical Understanding}:
Theoretical analysis of hypernetworks involves studying their representational capacity, learning dynamics, and generalization properties. By understanding the theoretical foundations of hypernetworks, researchers can gain insights into the underlying principles that drive their effectiveness and explore new avenues for improving their performance. Just like DNNs, understanding the working of hypernets is far from being solved. Although, there are some works that provide theoretical insights into hypernets, e.g., \citet{littwin2020infinite} highlighted that infinitely wide hypernetworks may not converge to a global minimum using gradient descent, but convexity can be achieved by increasing the dimensionality of the hypernetwork's output. \citet{Galanti2020} also studied the modularity of hypernets and showed that hypernets can be more efficient than the embedding-based method for mapping an input to a function. Intuitively, hypernets map an input to one point on a low-dimensional manifold for weights of target network \cite{shamsian2021personalized} -- theoretical insights into the connection between two can be very helpful. Thus, more research into the theoretical properties of hypernets will help to make them more popular and will also attract more research.

\textbf{Uncertainty-aware Deep Learning}:
Uncertainty-aware neural networks allow for more reliable and robust predictions, especially in scenarios where uncertainty estimation is crucial, such as decision-making under uncertainty, safety-critical applications, or when working with limited or noisy data \cite{abdar2021review}. Despite the success of DNNs and the development of different uncertainty quantification techniques, it still remains an open problem to quantify the prediction uncertainty \cite{kristiadi2019predictive}. Hypernets have opened a new door to uncertainty quantification as noise-conditioned hypernets can generate distribution on target network weights and have been shown to have better uncertainties than the SOTA \cite{krueger2018bayesian,ratzlaff2019hypergan}. Similarly, \citet{chauhan2023dynamic} used task-conditioned hypernets with dropout to generate multiple sets of weights for the target network. Further research into this can provide computationally efficient and effective techniques as compared with other techniques, such as ensemble methods, which need to train multiple models.

\textbf{Interpretability Enhancement}:
It will be helpful for the community to develop methods for visualizing, analyzing, and explaining the task-specific weights generated by hypernetworks. This includes developing intuitive visualization methods, and feature relevance analysis techniques that provide deeper insights into the weight generation and decision-making process of hypernetwork-based models.

\textbf{Model Compression and Efficiency}:
Hypernetworks can aid in model compression and efficiency in some problem settings \cite{zhao2020meta,mahabadi2021parameter}, where smaller hypernets are trained to generate larger target networks that can reduce the memory footprint and computational requirements of the model. This is particularly useful in resource-constrained environments where memory and computational resources are limited, and hypernets can be studied specifically for such settings.

\textbf{Usage Guidelines}:
Hypernetworks add additional complexity to solving problems. {\color{black}As with HyperDNN, we have an additional network to generate weights for the target DNN}. Hypernets introduce additional hyperparameters related to the weight generation process, e.g., what kind of weight generation should be used and how many chunks should be used. Some research and guidelines are needed to guide the researchers through these choices, stressing the need for a comparative study of different approaches under varying problem settings.

Thus, the field of hypernetworks in deep learning presents several challenges and opportunities for future research. The advancements in these areas will pave the way for the widespread adoption and effective utilization of hypernetworks in various domains of deep learning.

\section{Conclusion}
\label{sec_conclusion}
Hypernetworks have emerged as a promising approach to enhance deep learning models with increased flexibility, efficiency, generalization, uncertainty awareness, and information sharing. They have opened new avenues for research and applications across various domains. In this paper, we presented the first review of hypernetworks in the context of deep learning. We provided an illustrative example to explain the workings of hypernetworks and proposed a categorization based on five design criteria: inputs, outputs, variability of inputs and outputs, and the architecture of hypernets.
We discussed some of the important applications of hypernets to different deep learning problems, including multitasking, continual learning, federated learning, causal inference, and computer vision. Additionally, we presented scenarios and questions to help readers understand whether hypernets can be applied to a given problem setting. Finally, we highlighted challenges that need to be addressed in the future. These challenges include initialization, stability, scalability, efficiency, and the need for theoretical insights.
Future research should focus on tackling these challenges to further advance the field of hypernetworks and make them more accessible and practical for real-world applications. By addressing these issues, the potential of hypernetworks can be fully realized, leading to more robust and versatile deep learning models.

\section*{Acknowledgements}
This work was supported in part by the National Institute for Health Research (NIHR) Oxford Biomedical Research Centre (BRC) and in part by InnoHK Project Programme 3.2: Human Intelligence and AI Integration (HIAI) for the Prediction and Intervention of CVDs: Warning System at Hong Kong Centre for Cerebro-cardiovascular Health Engineering (COCHE).
DAC was supported by an NIHR Research Professorship, an RAEng Research Chair, the InnoHK Hong Kong Centre for Cerebro-cardiovascular Health Engineering (COCHE), the NIHR Oxford Biomedical Research Centre (BRC), and the Pandemic Sciences Institute at the University of Oxford. 
The views expressed are those of the authors and not necessarily those of the NHS, the NIHR, the Department of Health, the InnoHK – ITC, or the University of Oxford.


\section*{Statements and Declarations}
\subsection*{Competing Interests}
The authors declare that they have no competing interests.

\subsection*{Data Availability}
Data sharing not applicable to this article as no datasets were generated or analysed during the current study.

\subsection*{Contributions}
V.K.C. conceptualized the study, analyzed the literature and wrote the first draft. J.Z., P.L. and S.M. helped to filter out the literature and prepare Table~\ref{tab_hypernets}. D.A.C did supervision and funding acquisition. All authors reviewed and approved the final manuscript.

\bibliographystyle{apalike}
\bibliography{ML4HC}  




\end{document}